\documentclass[runningheads]{llncs}
\usepackage{graphicx}
\usepackage{subcaption}
\usepackage{booktabs}
\usepackage[hidelinks]{hyperref}

\usepackage{cite}
\usepackage{amsmath,amssymb,amsfonts}
\usepackage{algorithmic}
\usepackage{graphicx}
\usepackage{pdfpages}
\usepackage{multirow}
\usepackage{array}
\usepackage{caption}
\usepackage{float}

\usepackage[T5]{fontenc}
\usepackage{balance}
\usepackage[utf8]{inputenc}
\DeclareTextSymbolDefault{\DH}{T1}
\usepackage{lmodern} 

\usepackage{textcomp}
\usepackage{xcolor}
\def\BibTeX{{\rm B\kern-.05em{\sc i\kern-.025em b}\kern-.08em
    T\kern-.1667em\lower.7ex\hbox{E}\kern-.125emX}}

\usepackage{orcidlink}

\begin{document}

\title{CIAN: Multi-Stage Framework for Event-Enriched Image Captioning via Retrieval-Augmented Generation}

\titlerunning{CIAN: Multi-Stage Framework for Event-Enriched Image Captioning}

\author{%
Trinh Thi Thu Hien\inst{1,2}\orcidlink{0009-0006-0396-7219} \and Trung-Nghia Le\inst{1,2}\orcidlink{0000-0002-7363-2610}\thanks{Corresponding author.}
}

\institute{
\textsuperscript{1}Faculty of Information Technology, University of Science, Ho Chi Minh City\\
\textsuperscript{2}Vietnam National University, Ho Chi Minh City,  Vietnam\\
\email{23120254@student.hcmus.edu.vn}, \email{ltnghia@fit.hcmus.edu.vn}
}

\authorrunning{Trinh Thi Thu Hien and Trung-Nghia Le}

\maketitle

\begin{abstract}
Event-enriched image captioning describes not only visible content but also the broader context of events, including timing, location, and participants, capabilities missing in most pixel-bound models. We propose the Contextual Image-Article Narrator (CIAN), a multi-stage framework that enriches captions with external narratives. CIAN retrieves relevant articles using SigLIP, summarizes them to guide a Narrative Generation stage with a LoRA-fine-tuned Qwen model, and applies N-Gram-based Refinement for fluency and coherence. On the OpenEvents-V1 benchmark, CIAN achieves high retrieval performance (mAP 0.979) and improves caption quality, increasing CIDEr from 0.030 to 0.094. These results highlight the effectiveness of retrieval-augmented reasoning combined with linguistic refinement for generating context-aware, human-like captions.

\keywords{Event-enriched image captioning, Retrieval-augmented generation, Context retrieval, Narrative generation, LoRA fine-tuning.}
\end{abstract}

\section{Introduction}

Image captioning, at the intersection of computer vision and natural language processing, aims to generate natural textual descriptions of images~\cite{stefanini2021tellsurveydeeplearningbased}. The field has evolved from template-based and retrieval methods~\cite{NIPS2011_5dd9db5e, 10.1007/978-3-319-10593-2_35} to deep learning encoder–decoder architectures~\cite{sutskever2014sequencesequencelearningneural, kalchbrenner-blunsom-2013-recurrent}. More recently, Multimodal Large Language Models (MLLMs)~\cite{Yin_2024, sarto2025imagecaptioningevaluationage} produce semantically rich, paragraph-level narratives with sophisticated visual understanding.

Despite these advances, models remain pixel-bound and struggle to capture deeper contextual information, such as event significance or participant roles~\cite{eventa25}. Event-enriched image captioning requires reasoning beyond visual recognition, integrating spatio-temporal and external knowledge. Without retrieval mechanisms, captions often remain shallow~\cite{9849164}.

To address this, we propose the Contextual Image-Article Narrator (CIAN), a multi-stage framework that enriches captions with external, non-visual knowledge. CIAN retrieves semantically relevant articles using the SigLIP vision-language model~\cite{zhai2023sigmoidlosslanguageimage}, generates narratives by combining visual and textual information through storytelling-oriented prompts, and refines captions with an N-Gram-based mechanism for fluency and alignment with human references. Experiments on the OpenEvents-V1 benchmark~\cite{openeventsv1} show that CIAN achieves high retrieval accuracy (mAP 0.979, Recall@1 0.969) and improves caption quality (CIDEr score from 0.030 to 0.094), demonstrating its effectiveness in producing contextually grounded, human-like captions.

Our contributions as follows:  
\begin{itemize}
\item Introducing CIAN, a multi-stage framework combining retrieval, narrative generation, and linguistic refinement for context-enriched captions.  
\item Designing a Context Retrieval module using SigLIP to link images with relevant articles for accurate, event-specific information.  
\item Developing a Narrative Generation process that fuses visual cues with retrieved text through storytelling prompts for context-aware captions.  
\item Implementing N-Gram-based Refinement to enhance fluency and align generated captions with human-written stylistic patterns.  
\item Demonstrating through experiments that CIAN substantially improves retrieval and captioning performance on OpenEvents-V1.
\end{itemize}

\section{Related Work}

\textbf{Image-to-Text Retrieval.} Image-to-text retrieval connects visual content with textual descriptions, which is crucial for tasks such as content-based image search and knowledge-enriched captioning. Early methods projected images and text into a shared latent space using cross-modal embeddings. VSE++~\cite{faghri2018vseimprovingvisualsemanticembeddings} improved this with hard negative mining, and SCAN~\cite{lee2018stackedcrossattentionimagetext} used cross-attention for region–word alignment. Recent Vision-Language Pretrained Models such as CLIP~\cite{radford2021learningtransferablevisualmodels}, ALIGN~\cite{jia2021scalingvisualvisionlanguagerepresentation}, and BLIP~\cite{li2022blipbootstrappinglanguageimagepretraining} achieve strong cross-modal alignment through large-scale pretraining. Sparse retrieval relies on lexical matching, while dense retrieval~\cite{karpukhin2020densepassageretrievalopendomain} uses vector embeddings for semantic matching. Unlike prior work on single-sentence retrieval, we focus on document-level retrieval to capture richer context. Our approach leverages Retrieval-Augmented Generation~\cite{lewis2021retrievalaugmentedgenerationknowledgeintensivenlp} to ground captions in retrieved knowledge for more informative outputs.

\textbf{Image Captioning.} Image captioning generates textual descriptions from visual content~\cite{DBLP:journals/corr/abs-1810-04020}, evolving from encoder–decoder models to Transformer-based architectures with attention~\cite{DBLP:journals/corr/abs-1912-08226}, evaluated using BLEU\cite{papineni-etal-2002-bleu}, METEOR\cite{banerjee-lavie-2005-meteor}, and CIDEr~\cite{Vedantam_2015_CVPR}. While these models provide literal captions, they often lack context. Multimodal Large Language Models (MLLMs) enable richer, paragraph-level captions by leveraging world knowledge and reasoning. However, even advanced models may miss non-visual context such as identities, events, or significance. Retrieval-augmented and knowledge-based methods address this by integrating external sources, providing broader narrative context. Our work extends this by combining large-scale news retrieval with a fine-tuned language model to generate captions that are both descriptive and contextually informative.

\section{Proposed Method}

\subsection{Overview}

\begin{figure}[t!]
    \centering
    \includegraphics[width=\linewidth]{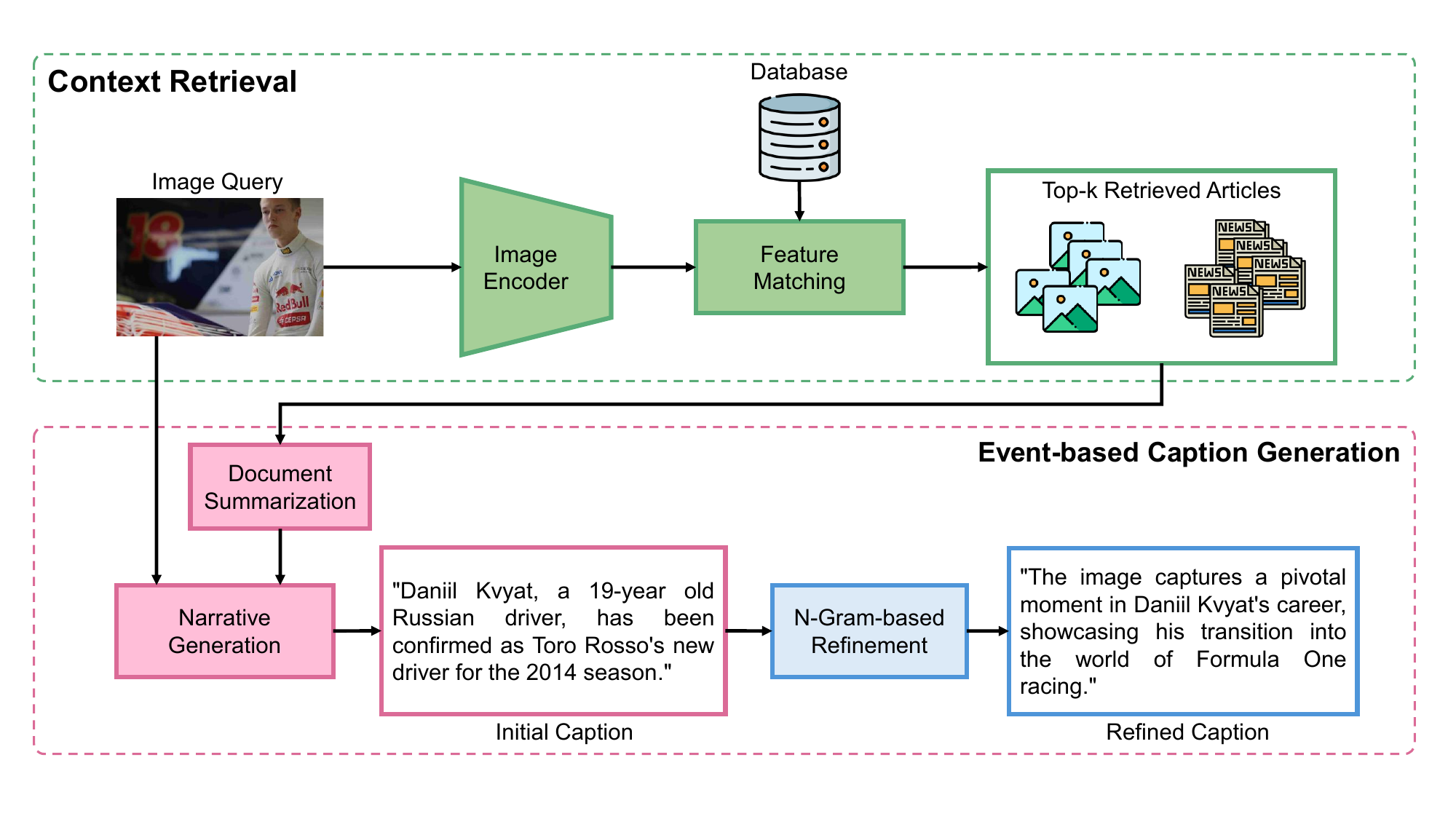}
    \caption{Overview of the proposed Contextual Image-Article Narrator (CIAN) framework. 
    The pipeline consists of three main stages: \textit{Context Retrieval}, which leverages the SigLIP model to retrieve semantically relevant articles for each query image; 
    \textit{Narrative Generation}, where a LoRA-fine-tuned Qwen model synthesizes event-aware captions by integrating visual and textual information; 
    and \textit{N-Gram-based Refinement}, which enhances linguistic fluency and domain consistency.}
    \label{fig:overview}
\end{figure}

To tackle the challenge of event-enriched image captioning, we propose a multi-stage pipeline that retrieves relevant contextual information and generates coherent, informative narratives. The framework consists of four main stages: (1) Dataset Enhancement through targeted web crawling to improve image–text alignment; (2) Context Retrieval using the SigLIP vision-language model to identify semantically relevant articles; (3) Narrative Generation with a prompt-engineered, LoRA-fine-tuned Qwen model guided by BART-based summarization; and (4) N-gram-based Refinement to enhance caption fluency and stylistic consistency by aligning with the linguistic distribution of the target domain.

\subsection{Dataset Curation and Enhancement}
A key limitation of the OpenEvents-V1 dataset \cite{openeventsv1} lies in its high signal-to-noise ratio caused by weak image–text alignment. In many cases, an image is only loosely related to its paired article, with relevant references buried deep within lengthy text. This misalignment poses a challenge for retrieval-augmented models, which may be misled by irrelevant or noisy content during caption generation.

To address this issue, we implemented a targeted data curation pipeline. Specifically, 3,323 articles contained “hide caption” markers, indicating that their associated images corresponded to only one or two sentences within the full article. For this subset, we developed a custom web crawler to bypass full-text parsing and directly access image galleries on CNN’s website. This process yielded approximately 23,000 high-quality images with explicit captions. We then integrated these image–caption pairs into the dataset by using the SigLIP model to match each with its most visually similar image from the original collection, thereby improving the overall alignment quality.
\subsection{Context Retrieval}

The retrieval phase aims to identify the most semantically relevant articles for a given query image by leveraging the image--text alignment capabilities of the SigLIP model~\cite{zhai2023sigmoidlosslanguageimage}. This process includes two main steps: image-to-image retrieval and text embedding aggregation for article scoring.

\subsubsection{Image Query Embedding and Retrieval}

Given a query image $I_q$, its embedding $\mathbf{e}_q$ is obtained using the SigLIP image encoder. Each image in the database $\{I_i\}_{i=1}^N$ has a pre-computed embedding $\mathbf{e}_i$. The similarity between the query and each database image is measured using cosine similarity, defined as the normalized dot product between $\mathbf{e}_q$ and $\mathbf{e}_i$. The top-$K$ most similar images (for example, $K = 20$) are selected based on descending similarity scores, forming the retrieved index set $\mathcal{R} = \{i_1, i_2, \dots, i_K\}$.

\subsubsection{Article Scoring via Text Embedding Aggregation}

Each retrieved image $I_{i_k}$ corresponds to a candidate article $A_{i_k}$ that includes a title $T_{i_k}$ and body content $C_{i_k}$. Both components are encoded using the SigLIP text encoder to produce embeddings $\mathbf{e}^{t}_{i_k}$ and $\mathbf{e}^{c}_{i_k}$. A weighted combination of these embeddings, where $\lambda = 0.3$, is used to create the final article representation: $\mathbf{e}^{article}_{i_k} = \lambda \mathbf{e}^{t}_{i_k} + (1 - \lambda)\mathbf{e}^{c}_{i_k}$. The relevance between the query image and each article is then computed using cosine similarity between $\mathbf{e}_q$ and $\mathbf{e}^{article}_{i_k}$. Articles are finally re-ranked according to these scores, and the highest-ranked entries are selected as the final retrieval results.

\subsection{Narrative Generation}

The core of our captioning system is a generative module built on the Qwen2.5-VL model \cite{Qwen2VL}, fine-tuned using the parameter-efficient Low-Rank Adaptation (LoRA) technique \cite{hu2022lora}. To equip the model with concise and relevant event context, we summarize each accompanying news article using a pre-trained BART model \cite{lewis2019bart}, extracting the most salient information while reducing textual redundancy. During inference, the fine-tuned LoRA model takes as input the query image and the summarized article, which are integrated into a storytelling-oriented prompt. Guided by this structured prompt, the model generates an initial event-aware caption, denoted as 
$c_{\text{story}}$, that captures both the visual and contextual aspects of the event. This caption serves as the output of the generative stage and the input for the subsequent refinement stage.

\subsection{N-Gram-based Refinement}

Recognizing that the CIDEr metric~\cite{vedantam2015ciderconsensusbasedimagedescription} is highly sensitive to n-gram overlap with ground-truth captions, we introduce a caption refinement stage explicitly designed to enhance this alignment. The objective of this module is to improve the lexical and structural similarity between generated captions and human-written references, thereby boosting reference-based metric performance without compromising semantic fidelity.

Let $\mathcal{D} = \{y_1, y_2, \dots, y_N\}$ denote the set of ground-truth captions in the training set. We compute frequency distributions of unigrams ($n=1$), bigrams ($n=2$), and trigrams ($n=3$) across $\mathcal{D}$. The set of high-frequency n-grams $\mathcal{G}_n$ includes those with frequency above a threshold $\tau_n$ for each n-gram length, and the final n-gram lexicon $\mathcal{G}$ is obtained by uniting all $\mathcal{G}_n$ for $n \in \{1, 2, 3\}$.

Given an initial caption $c_{\text{story}}$ generated in the Narrative Generation stage, a refinement phase is introduced using a dedicated refiner prompt that encourages the inclusion of relevant terms from $\mathcal{G}$. The refined caption $c_{\text{final}}$ is produced through the QwenRefine model, which performs controlled rewriting to incorporate high-frequency n-grams in a contextually appropriate and semantically coherent manner. This process aligns the caption more closely with evaluation metrics such as CIDEr while maintaining fluency and narrative quality. The refinement strategy functions as a soft constraint that promotes n-gram usage without enforcing it rigidly, thereby avoiding unnatural phrasing. As demonstrated in our ablation study, this n-gram-driven augmentation improves CIDEr scores, validating the effectiveness of the proposed refinement mechanism.

\begin{figure}[t!]
    \centering
    \renewcommand{\arraystretch}{1.4}
    \begin{tabular}{@{} m{0.25\textwidth} m{0.72\textwidth} @{}}
        \includegraphics[width=\linewidth]{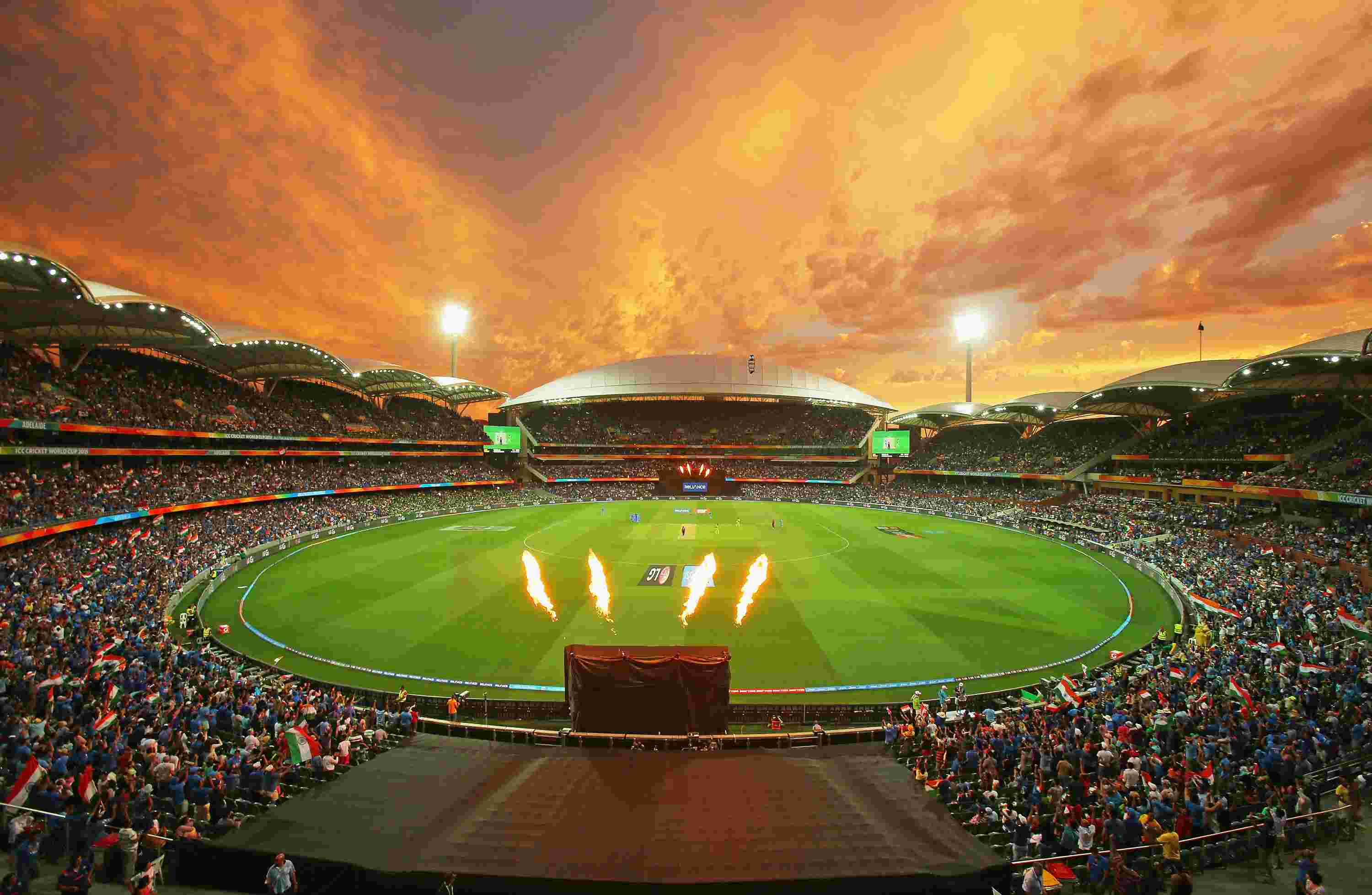} & 
        {\tiny This photograph captures a pivotal moment during the 2015 ICC Cricket World Cup match between arch-rivals India and Pakistan at Adelaide Oval. The stadium is awash with approximately one billion viewers, making it the most-watched cricket match in history. The image, despite not showing the players, focuses on the intensity of this event with a packed crowd and the prominent umpire's chair symbolizing the highly anticipated contest. The image's distorted perspective and vibrant colors emphasize the electric atmosphere, reflecting the global passion for cricket and the unique cultural significance it holds for India and Pakistan.} \\
    \end{tabular}
    \caption{Examples from the OpenEvents-V1 dataset with corresponding event-enriched captions.}
    \label{tab:dataset_examples}
\end{figure}

\section{Experiments}

\subsection{Implementation Details}
All experiments were conducted on Kaggle and Google Colab platforms using NVIDIA Tesla P100 and T4 GPUs. Our implementation utilized a PyTorch-based stack, with the fine-tuning process orchestrated by LLaMA Factory \cite{zheng2024llamafactory}.

We fine-tuned the Qwen2.5-VL-3B-Instruct model \cite{Qwen2VL} for 3 epochs using LoRA technique. Key hyperparameter included a LoRA rank of 4 applied to all linear layers, a LoRA+ learning rate ratio of 8.0, and an initial learning rate of 1e-4 with a cosine scheduler. To ensure training stability and efficiency, we employed an effective batch size of 4 (1 per device with 4 gradient accumulation steps), FP16 precision, and gradient clipping at a maximum norm of 1.0. The resulting LoRA adapters were subsequently merged into the base model for inference.


\subsection{Experimental Settings}
Our experiments were conducted on the public-test of the OpenEvents-V1 benchmark \cite{openeventsv1}, a large-scale dataset for multimodal event grounding that links 415,309 images with 202,803 news articles. Each article record includes metadata such as title, content, timestamp, and associated image identifiers, providing a rich yet often sparse alignment between visual scenes and textual event narratives (Fig.~\ref{tab:dataset_examples}).




To evaluate our proposed method, we followed the same evaluation pipeline as in the Eventa Track 1 challenge \cite{eventa25}, which assesses the dual tasks of retrieval and captioning. Methods were evaluated using the \textit{Overall Score}, which is a weighted harmonic mean of Average Precision (AP), Recall@1 (R@1), and Recall@10 (R@10), CLIPScore \cite{hessel2022clipscorereferencefreeevaluationmetric}, and CIDEr \cite{vedantam2015ciderconsensusbasedimagedescription}.

\begin{figure}[t!]
    \centering
    \renewcommand{\arraystretch}{1.4}
    \begin{tabular}{@{} m{0.25\textwidth} m{0.72\textwidth} @{}}
        \includegraphics[width=\linewidth]{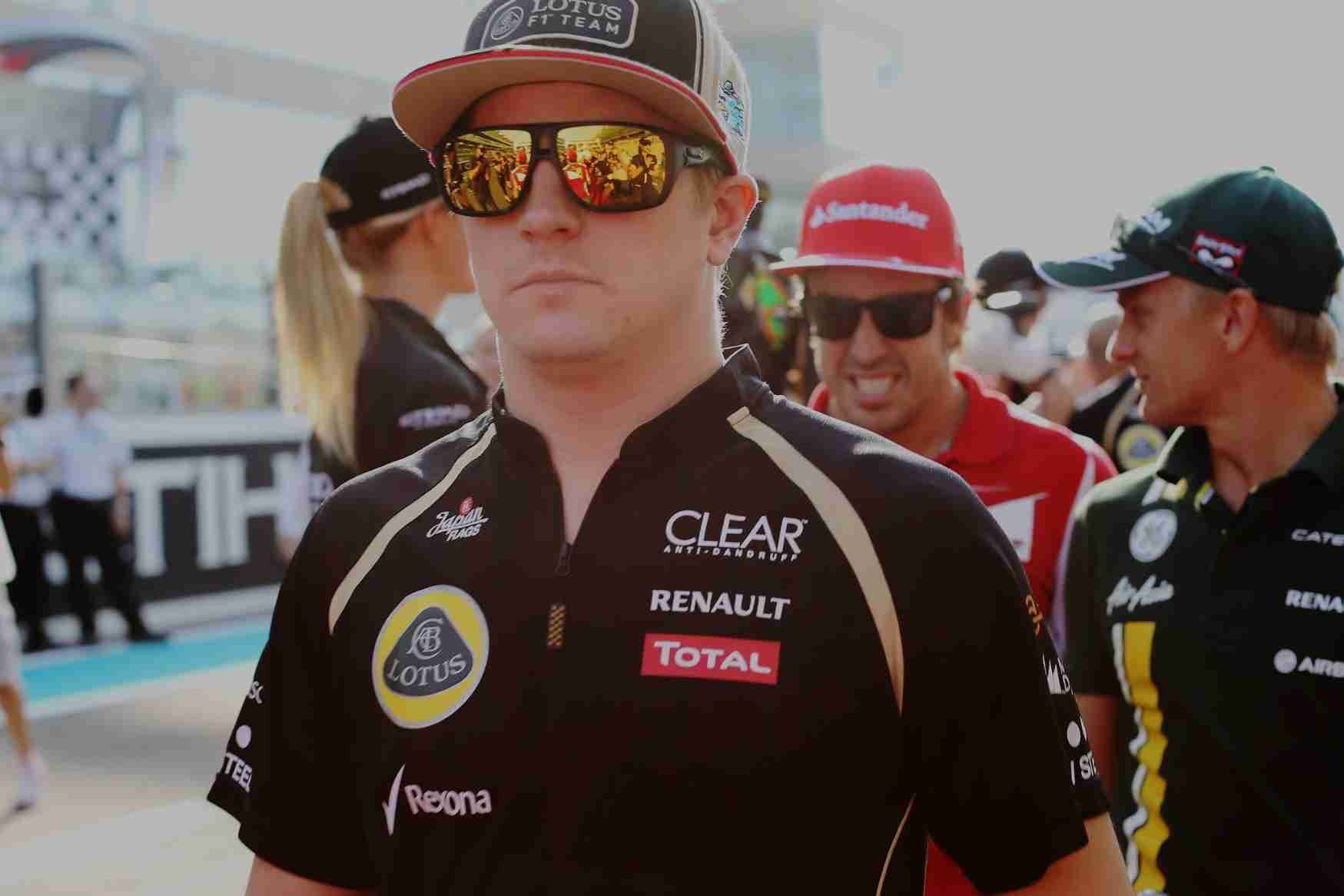} & 
       {\tiny This striking image captures a pivotal moment in Kimi Raikkonen's career, showcasing his confident demeanor amidst the bustling environment of the Lotus F1 Team paddock. Dressed in a black racing suit adorned with the team's vibrant yellow and blue accents, Raikkonen stands out against the backdrop of other team members, some wearing matching outfits while others don distinct attire. His sunglasses reflect the bright lights of the paddock, adding a sense of mystery and intrigue to his expression. The image, though slightly blurred, emphasizes Raikkonen's presence, suggesting a moment of anticipation or preparation for the upcoming race season. The Lotus logo prominently displayed on his suit reinforces his affiliation with the team and hints at the excitement surrounding the new 2013 car. While the specific event is not clear, the image undoubtedly captures a significant moment in Raikkonen's journey with Lotus, reflecting both his personal confidence and the team's aspirations for success in the competitive world of Formula One.} \\
        
    \end{tabular}
    \caption{Qualitative examples generated by the proposed framework.}
    \label{tab:result_examples}
\end{figure}

\subsection{Overall Results}

Figure \ref{tab:result_examples} shows qualitative examples from our framework, highlighting its ability to generate coherent, contextually enriched event descriptions that combine visual cues with retrieved textual knowledge.  

Our framework achieved an Overall Score of 0.332, demonstrating strong performance across retrieval and captioning tasks. The retrieval module achieved near-perfect accuracy with mAP 0.979, Recall@1 0.969, and Recall@10 0.996, confirming the effectiveness of the SigLIP-based strategy in identifying relevant contextual documents. For captioning, the model obtained a CLIP Score of 0.820 and a CIDEr score of 0.094, indicating strong semantic alignment and the benefit of prompt engineering and n-gram refinement. \textit{As no prior work exists for this benchmark, we report absolute metrics without comparisons.}



\subsection{Ablation Study}

\subsubsection{Impact of Proposed Components.}

To assess the impact of each generation stage, we conducted an ablation study on the captioning component using three progressively refined model configurations. As shown in Table~\ref{tab:ablation}, each stage contributed measurable improvements. The baseline model, which used a simple prompt for caption generation from the query image and retrieved article, achieved a CLIP score of 0.79 and a CIDEr of 0.014, indicating limited multimodal integration. Integrated Dataset Enhancement, which replaced misaligned pairs through targeted re-crawling and SigLIP-based alignment, the CLIP score remained unchanged while the CIDEr increased to 0.030. Consequently, the Narrative Generation shows effectiveness in improving CIDEr to 0.076 but reducing the CLIP score slightly to 0.76. Finally, incorporating N-gram-based Refinement enhanced linguistic fluency and contextual alignment, yielding the best results with a CLIP score of 0.82 and a CIDEr of 0.094, confirming the effectiveness of progressive refinement in boosting caption quality.

\begin{table}[t!]
\centering
\caption{Ablation study on the caption generation process. Each stage shows a significant and cumulative improvement. Qwen, DE, NG, and NR denote Re-ranking, Qwen2.5-VL-3B-Instruct, Dataset Enhancement, Narrative Generation, and N-gram Refinement, respectively.}
\label{tab:ablation}
\begin{tabular}{l|cccc|l|l}
\toprule
\multicolumn{1}{c|}{\multirow{2}{*}{\textbf{Model}}} & \multicolumn{4}{c|}{\textbf{Configurations}} & \multicolumn{2}{c}{\textbf{Metrics}} \\
\cmidrule{2-7}
\multicolumn{1}{c|}{} & \textbf{Qwen} & \textbf{DE} & \textbf{NG} & \textbf{NR} & \multicolumn{1}{|c}{\textbf{CLIP Score}} & \multicolumn{1}{|c}{\textbf{CIDEr Score}} \\
\midrule
Baseline 1 & x &  &  &  & 0.79 & 0.014 \\
Baseline 2 & x & x &  &  & 0.79 (+0.00) & 0.030 (+0.016)\\
Baseline 3 & x & x & x &  & 0.76 (-0.03) & 0.076 (+0.046) \\
CIAN (Ours) & x & x & x & x & \textbf{0.82} (+0.06) &  \textbf{0.094 }(+0.018)
\\
\bottomrule
\end{tabular}
\end{table}


\subsubsection{Effectiveness of Clip-based Metrics}

The selection of an effective vision encoder is critical for robust visual understanding and accurate image retrieval. To validate our choice of SigLIP, we conduct a comparative analysis against several pre-trained vision models, including CLIP, BLIP, and DINOv2. As shown in Table~\ref{tab:study_text_encoder}, SigLIP achieved the highest results across all metrics, with an mAP of 0.979, Recall@1 of 0.969, and Recall@10 of 0.996. This empirical evidence confirms that SigLIP produces more robust and discriminative image representations for our task. Consequently, we adopt it as the primary vision encoder in our system.



\begin{table}[t!]
\centering
\caption{Performance of different feature encoders on the image retrieval task.}
\label{tab:study_text_encoder}
\begin{tabular}{l|ccc}
\toprule
\textbf{Method} & \textbf{mAP} & \textbf{Recall@1} & \textbf{Recall@10} \\ \midrule
CLIP & 0.971 & 0.958 & 0.990 \\
BLIP & 0.940 & 0.914 & 0.985 \\
DinoV2 & 0.966 & 0.950 & 0.989 \\ 
\textbf{SigLIP} & \textbf{0.979} & \textbf{0.969} & \textbf{0.996} \\
\bottomrule
\end{tabular}
\end{table}



\subsection{Failure Cases and Discussions}

\begin{figure}[t!]
\centering
\begin{tabular}{cccc}
\textbf{Query Image} & \multicolumn{3}{c}{\textbf{Retrieved Images}} \\ \cmidrule(lr){2-4}
 & \textbf{Top 1} & \textbf{Top 2} & \textbf{Top 3} \\

\includegraphics[width=0.2\linewidth]{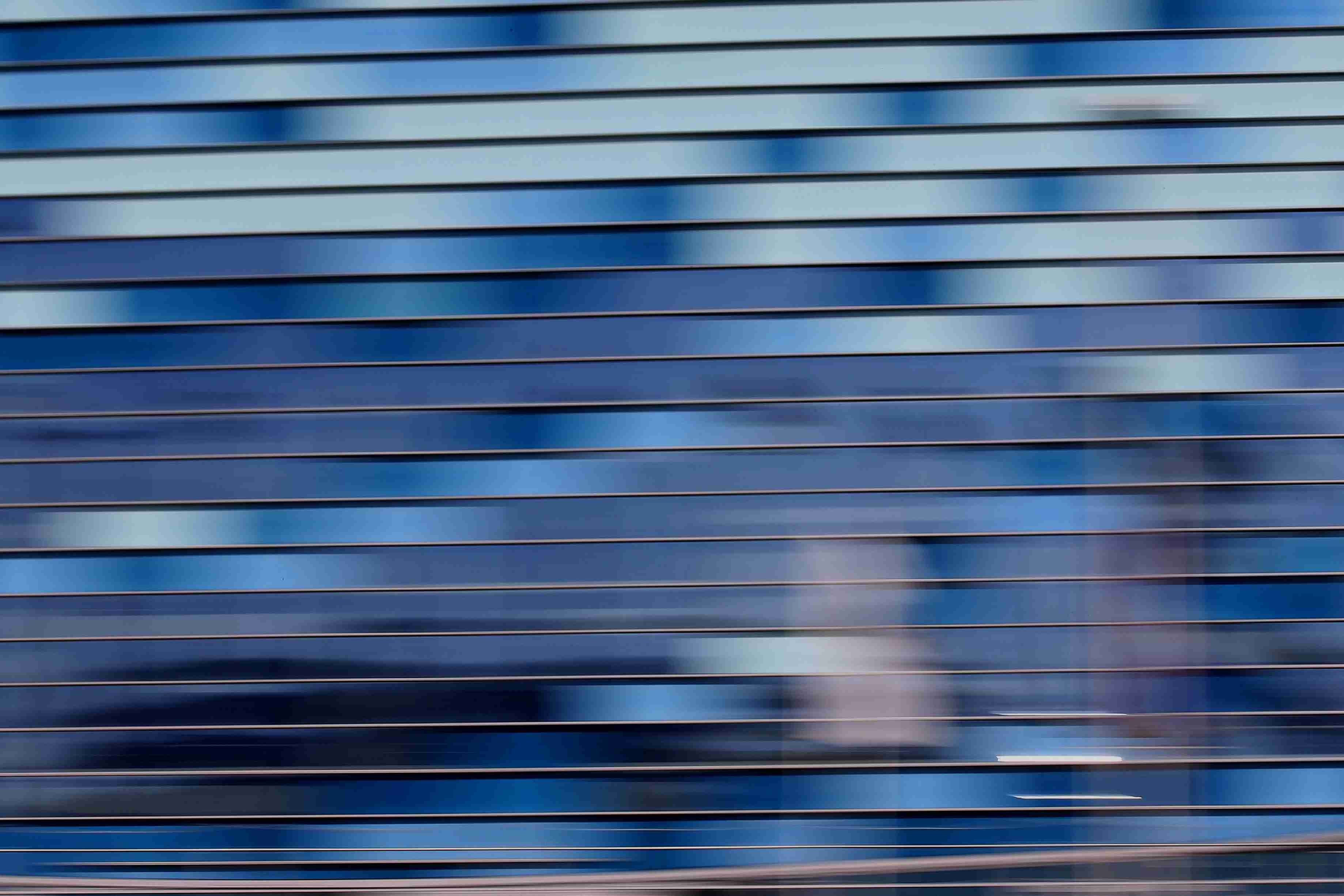} &
\includegraphics[width=0.2\linewidth]{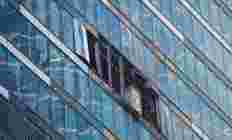} &
\includegraphics[width=0.2\linewidth]{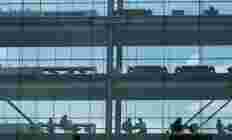} &
\includegraphics[width=0.2\linewidth]{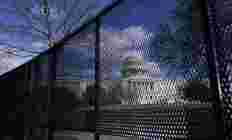} \\

\includegraphics[width=0.2\linewidth]{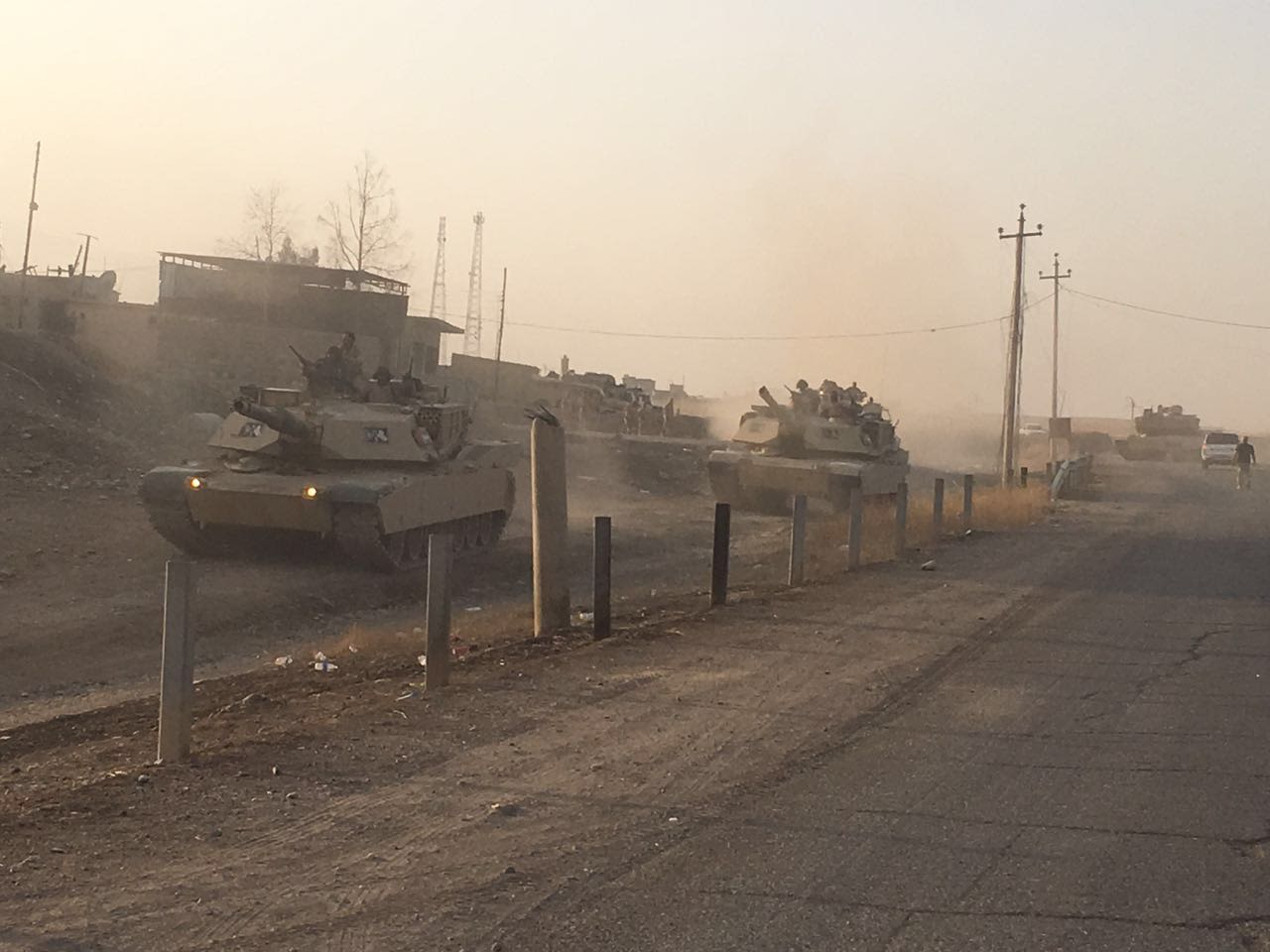} &
\includegraphics[width=0.2\linewidth]{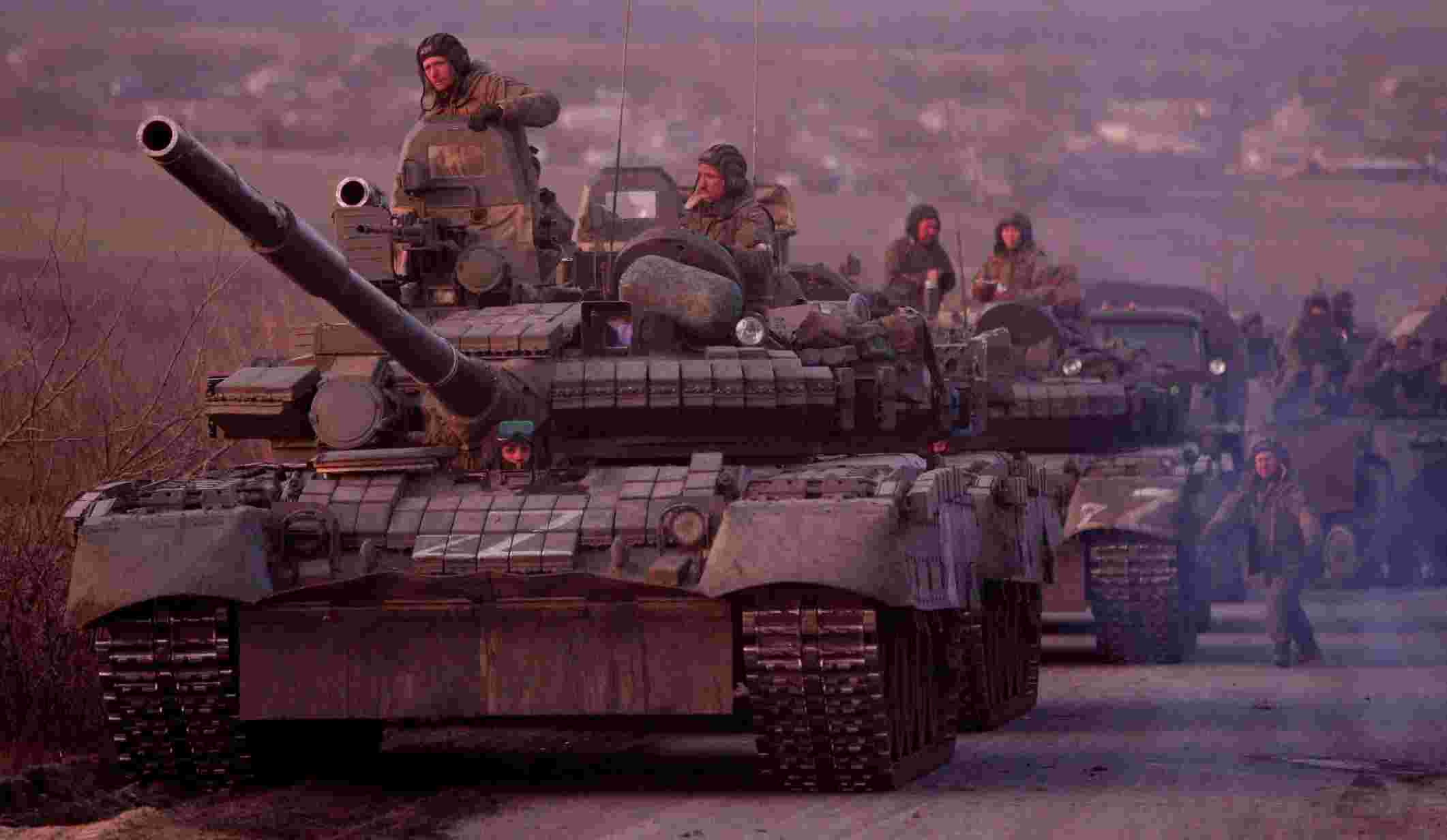} &
\includegraphics[width=0.2\linewidth]{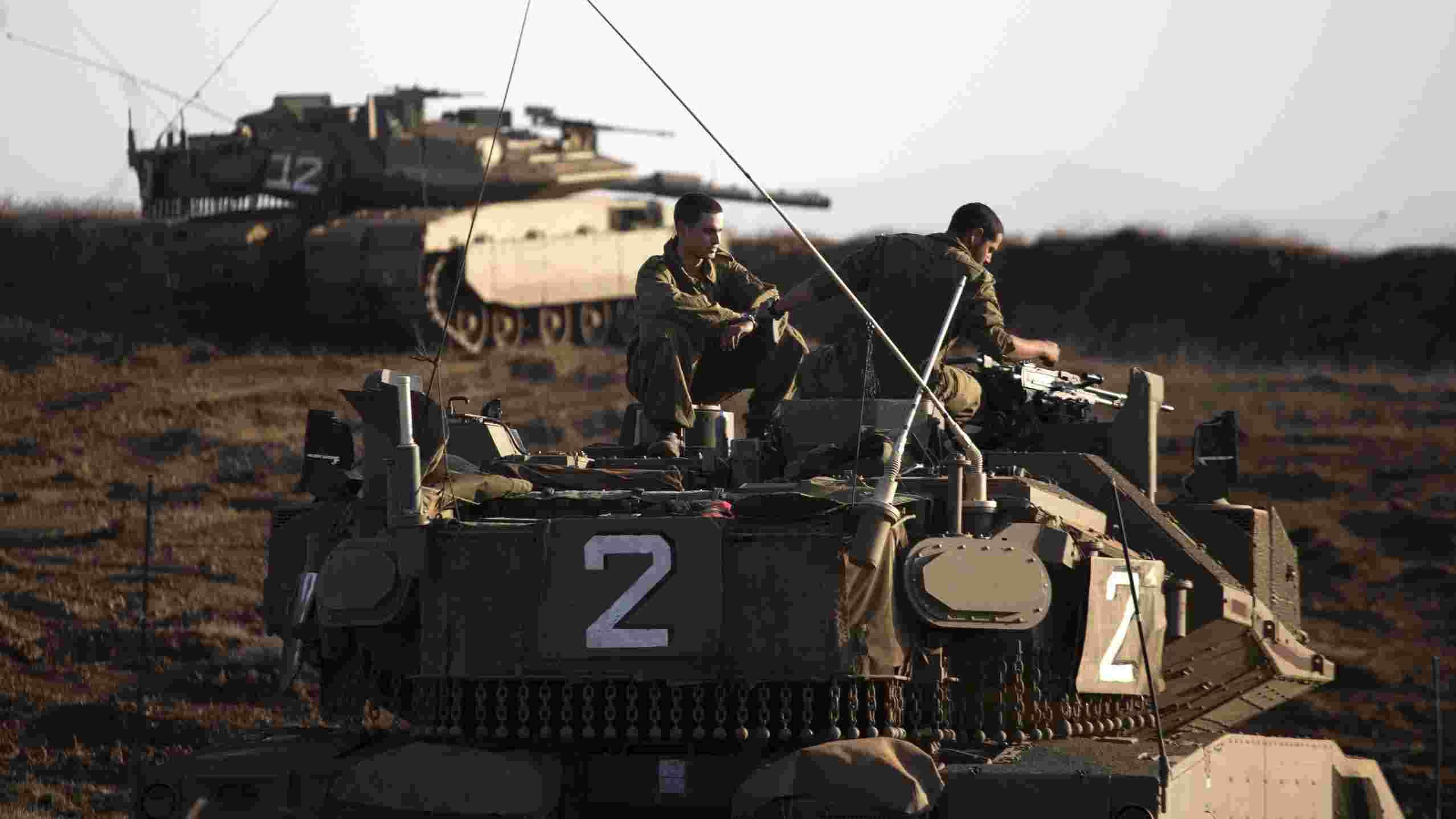} &
\includegraphics[width=0.2\linewidth]{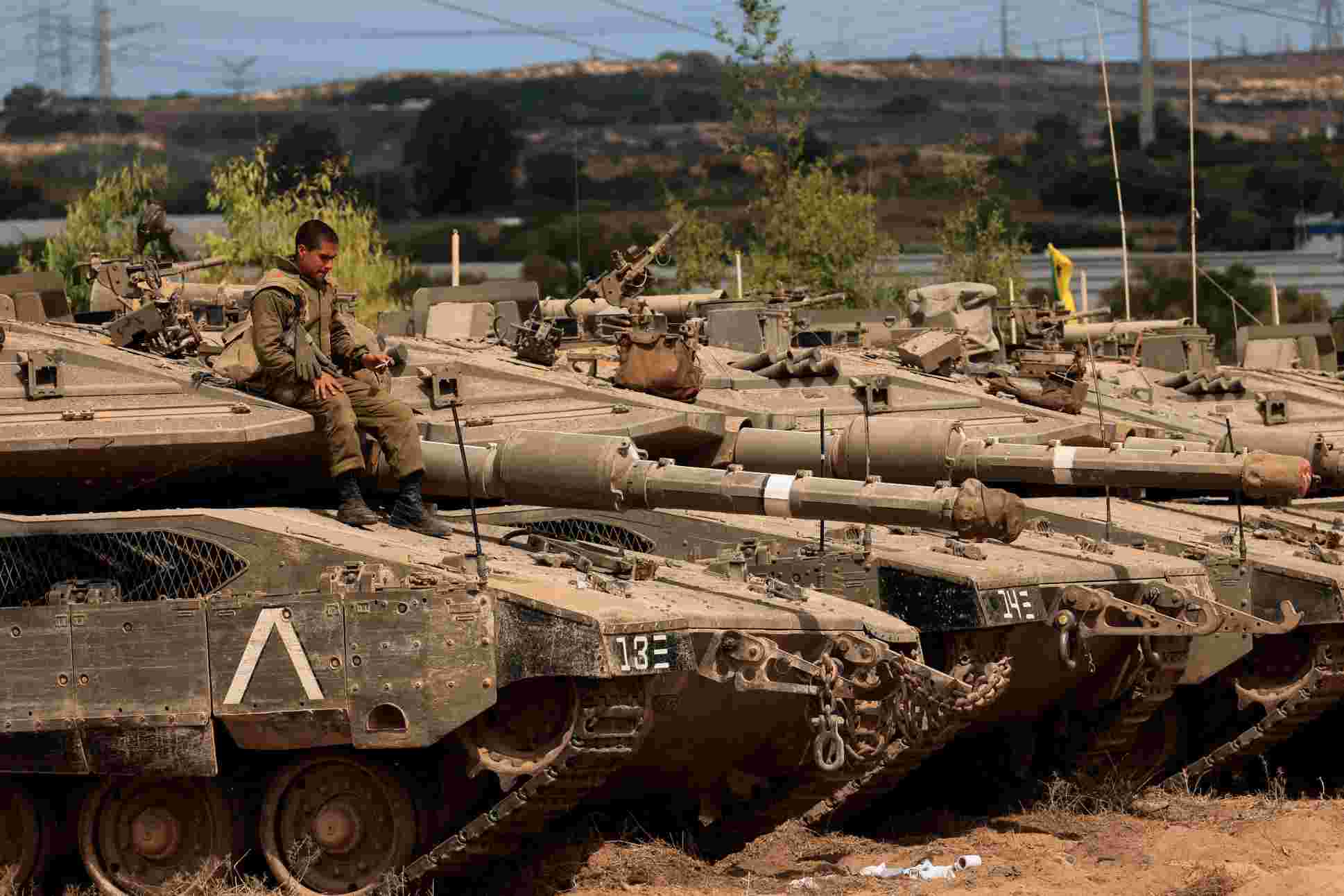} \\
\end{tabular}
\caption{Failure cases of the retrieval results produced by our framework.}
\label{fig:failure1}
\end{figure}

\subsubsection{Incorrect Image-to-Image Retrieval.}
The mismatch between the query image and the retrieved result can be attributed to limitations in the dataset quality (Fig.\ref{fig:failure1}). Using a scaled-down version of the dataset, while efficient in terms of computational cost and training time, introduces a trade-off that reduces retrieval accuracy. Additionally, the model struggles to retrieve the correct image when the query image is subjected to visual augmentations.

\subsubsection{Entity Grounding and Narrative Integration Error.}

\begin{figure}[t!]
    \centering
    \renewcommand{\arraystretch}{1.4}
    \begin{tabular}{@{} m{0.25\textwidth} m{0.72\textwidth} @{}}
        \includegraphics[width=\linewidth]{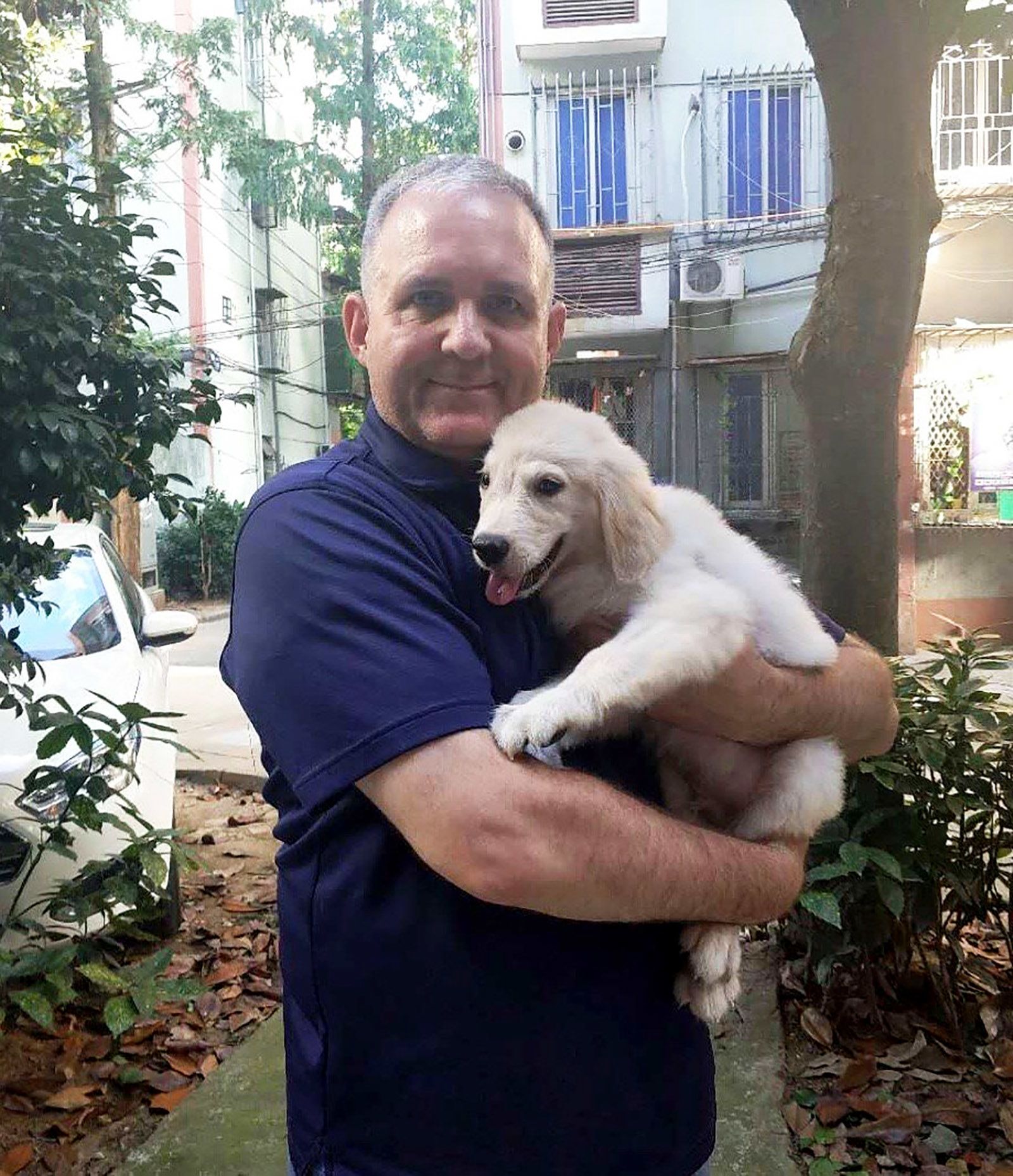} & 
        {\tiny "A man in a blue shirt cradles a small white puppy in his arms, standing outdoors against a backdrop of a building."}
    \end{tabular}
    \caption{Our method accurately captures the visual content but fails to recognize the person as Paul Whelan or connect the image to the article’s narrative about his detention.}
    \label{fig:failure2}
\end{figure}

A recurring failure pattern arises from weak alignment between visual subjects and the corresponding named entities or narrative context. As illustrated in Fig.~\ref{fig:failure2}, the model can describe visual details accurately but often fails to identify individuals or integrate textual context meaningfully. This reflects two main issues: the lack of entity grounding, preventing visual–textual linking, and poor narrative fusion, where textual information is added rather than contextually integrated. These limitations suggest shallow cross-modal correlations, which future work could address through multimodal entity linking, cross-attentional fine-tuning, and coherence-based reinforcement learning.

\subsubsection{Ineffectiveness of Dataset Enhancement}

The current data crawling algorithm, which activates only when a “hide caption” marker is detected, is not fully effective. Many mismatched image–article pairs cannot be identified using this criterion. As shown in Fig.~\ref{fig:failure3}, some images have captions that misalign with their article content but lack the marker, leaving such errors undetected. 

\begin{table}[t!]
    \centering
    \caption{Our method provides an accurate visual description of the scene but fails to capture the main subject and contextual information from the newspaper.}
    \renewcommand{\arraystretch}{1.4}
    \begin{tabular}{@{} m{0.25\textwidth} | m{0.34\textwidth} | m{0.38\textwidth} @{}}
    \toprule
        \textbf{Image} & \textbf{Image Caption} & \textbf{Generated Caption} \\ \midrule
        \includegraphics[width=\linewidth]{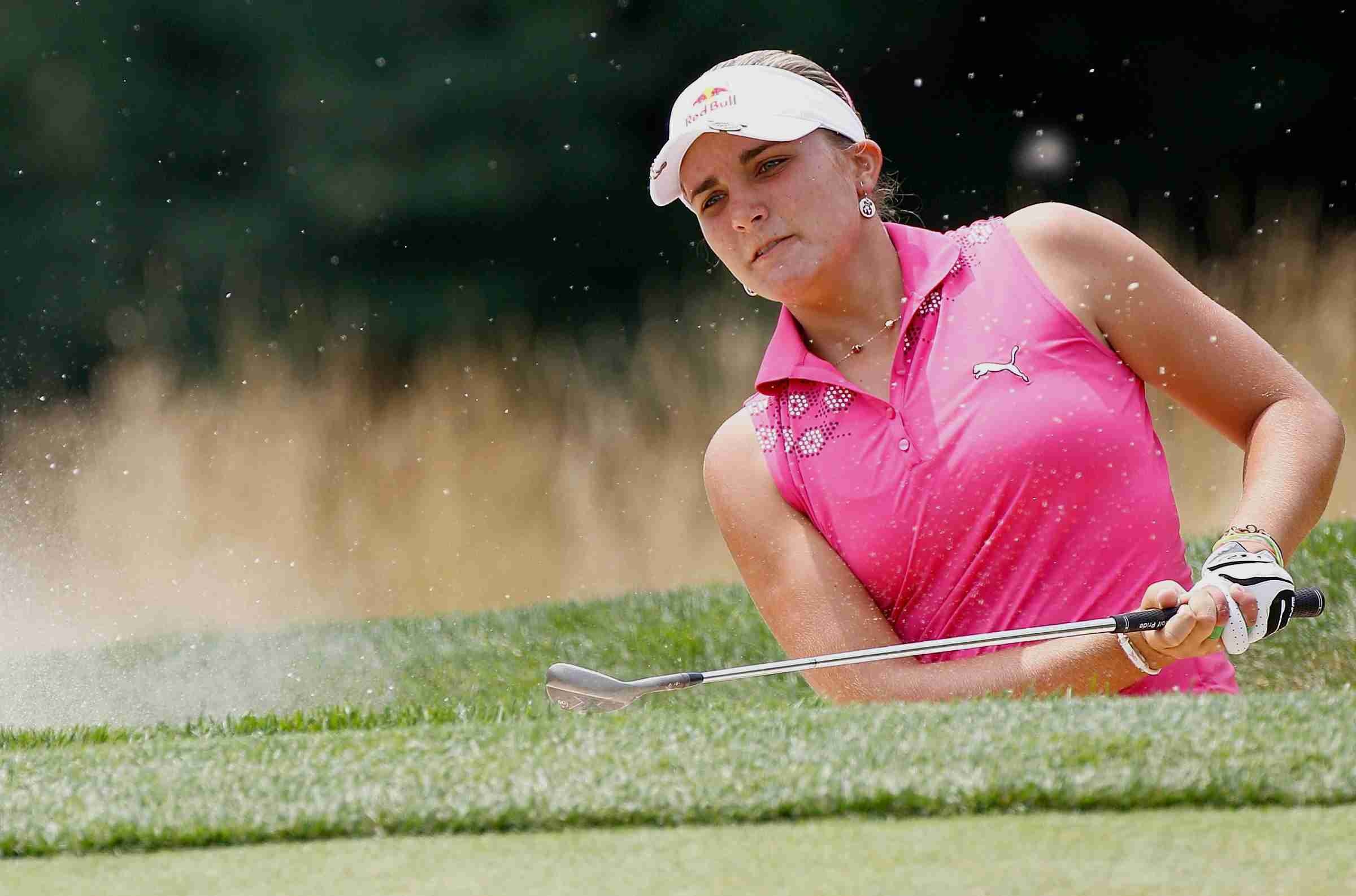} & 
        {\tiny In June, Thompson announced she would be turning professional, just in time for her fourth consecutive US Women's Open. Thompson recorded her best ever finish at the tournament, tying for 10th and scooping her first professional paycheck -- a cool \$72,131.} & {\tiny A woman in a pink shirt and white visor is playing golf. She is hitting a golf ball out of a sand trap, creating a cloud of sand behind her. The image captures a moment of intense concentration and skill as she navigates the challenging terrain. The vibrant colors and dynamic composition highlight the excitement and athleticism of the sport.} \\
        \bottomrule
    \end{tabular}
    
    \label{fig:failure3}
\end{table}

\section{Conclusion}

We presented an optimized fine-tuning of the Qwen2.5-VL model for generating event-enriched image captions. Our method utilized dataset refinement and prompt engineering to boost contextual accuracy, achieving results with a CIDEr score of 0.094 and a CLIP score of 0.82. These findings confirm the value of progressive refinement in closing the semantic gap between visual content and external narratives.

Future work should investigate more advanced multimodal fusion mechanisms within MLLM architectures and also explore how different alignment models influence semantic and stylistic diversity to mitigate dataset biases and strengthen contextual understanding.

\section*{Acknowledgments}


This research used the GPUs provided by the Intelligent Systems Lab at the Faculty of Information Technology, University of Science, VNU-HCM. 

Trung-Nghia Le was funded by the Postdoctoral Scholarship Programme of Vingroup Innovation Foundation (VINIF), VinUniversity, code VINIF.2025.STS.14.

\bibliographystyle{splncs04}
\balance
\bibliography{references}

\end{document}